\documentclass[sigconf,nonacm]{acmart}
\AtBeginDocument{%
  }

\setcopyright{none}
\copyrightyear{2026}
\acmYear{2026}
\acmDOI{}
\acmConference[Preprint]{Technical report (preparation)}{}{}{}

\acmSubmissionID{}

\usepackage{booktabs}  
\usepackage{tabularx}  
\usepackage{array}
\usepackage{amsmath}

\usepackage{amssymb}
\newcolumntype{Y}{>{\centering\arraybackslash}X}
\usepackage{colortbl}  
\usepackage{bm}
\usepackage{graphicx}
\usepackage{subcaption}
\usepackage{float}
\usepackage[linesnumbered,ruled,vlined]{algorithm2e}
\usepackage[dvipsnames]{xcolor}
\usepackage[table]{xcolor}
\definecolor{bg_blue}{RGB}{228, 240, 255}
\definecolor{rose_red}{RGB}{219, 48, 122} 
\definecolor{bg_gray}{RGB}{240, 240, 240}
\definecolor{bg_pink}{RGB}{255, 240, 245} 
\definecolor{dark_bg_blue}{RGB}{153, 204, 255} 
\definecolor{bg1_blue}{RGB}{204, 229, 255}
\definecolor{cite_blue}{RGB}{70, 105, 170}

\newif\ifHamBRIncludeAppendix
\HamBRIncludeAppendixfalse

\begin{document}

\title{HamBR: Active Decision Boundary Restoration Based on Hamiltonian Dynamics for Learning with Noisy Labels}
\author{Ningkang Peng}
\affiliation{%
  \institution{Nanjing Normal University}
  \city{Nanjing}
  \country{China}}
\author{Jingyang Mao}
\affiliation{%
  \institution{Nanjing Normal University}
  \city{Nanjing}
  \country{China}}
\author{Qianfeng Yu}
\affiliation{%
  \institution{Nanjing Normal University}
  \city{Nanjing}
  \country{China}}
\author{Xiaoqian Peng}
\affiliation{%
  \institution{Nanjing University of Chinese Medicine}
  \city{Nanjing}
  \country{China}}
\author{Peirong Ma}
\affiliation{%
  \institution{Nanjing Normal University}
  \city{Nanjing}
  \country{China}}
\author{Yanhui Gu}
\affiliation{%
  \institution{Nanjing Normal University}
  \city{Nanjing}
  \country{China}}








\renewcommand{\shortauthors}{Peng et al.}

\begin{abstract}
    In large-scale visual recognition and data mining tasks, the presence of noisy labels severely undermines the generalization capability of deep neural networks (DNNs). Prevalent sample selection methods rely primarily on training loss or prediction confidence for passive screening. However, within a feature space degraded by noise, decision boundaries undergo systematic boundary collapse. This phenomenon hinders the ability of the model to distinguish between hard clean samples and noisy samples at the decision margins, thereby creating a significant performance bottleneck. This study is the first to emphasize the pivotal importance of active boundary restoration for noise-robust learning. We propose HamBR, a novel paradigm based on Hamiltonian dynamics. The core approach leverages the Spherical Hamiltonian Monte Carlo (Spherical HMC) mechanism to actively probe inter-class ambiguous regions within the representation space and synthesize high-quality virtual outliers. By imposing explicit repulsion constraints via energy-based modeling, these synthesized samples establish robust energy barriers at the decision boundaries. This mechanism forces real samples to move from dispersed overlapping regions toward their respective class centers, thereby restoring the discriminative sharpness of the decision boundaries. HamBR demonstrates exceptional versatility and can be integrated as a plug-and-play defense module into existing semi-supervised noisy label learning frameworks. Empirical evaluations show that the proposed paradigm significantly enhances the discriminative accuracy of hard boundary samples, achieving state-of-the-art (SOTA) performance on CIFAR-10/100 and real-world noise benchmarks. Furthermore, it exhibits superior convergence efficiency and reliable robustness, while improving significantly the capability of the model for Out-of-Distribution (OOD) detection.
    \end{abstract}

\maketitle


\section{Introduction}
In recent years, Deep Neural Networks (DNNs) have made significant progress in computer vision and data mining tasks \cite{medical2,zhou2025perception}. However, these achievements depend heavily on the availability of large-scale and high-quality annotated data. In real-world applications, the presence of label noise is nearly inevitable due to annotation costs or subjective bias \cite{song2025survey}. Training with noisy labels not only induces the model to overfit erroneous annotations but also significantly impairs the generalization performance of the model \cite{performence3}.

\begin{figure}[t]  
    \centering
    \includegraphics[width=1\linewidth]{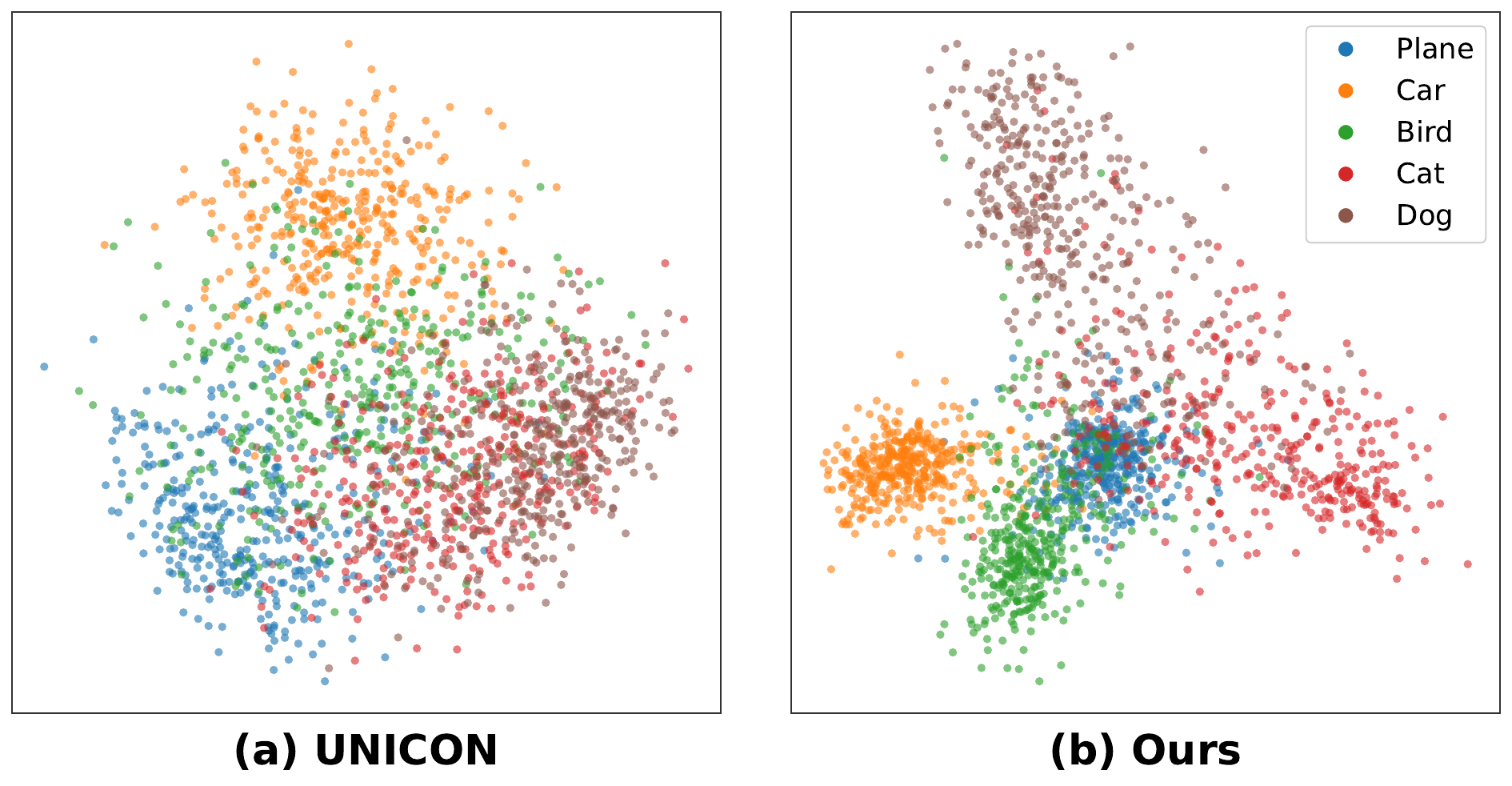} 
    \caption{\textbf{Comparison of feature distributions.} (a) UNICON suffers from severe boundary collapse, where distinct classes are entangled in the latent space, making it difficult to distinguish hard samples. (b) Ours (HamBR) successfully achieves active boundary rectification by synthesizing virtual outliers and enforcing geometric constraints, significantly enhancing inter-class separability.}
    \label{fig:tu}
\end{figure}

Existing research is primarily divided into two categories: noise transition matrix estimation \cite{patrini2017making,goldberger2017training} and sample selection \cite{Han9}. The paradigm of sample selection has become mainstream due to its scalability, typically using the small-loss criterion or prediction confidence to filter clean samples \cite{li2020dividemix,arpit2017closer}. However, these passive screening mechanisms face a critical bottleneck: label noise systematically disrupts the geometric structure of the feature space, leading to boundary collapse \cite{bai2021understanding,Sel15}, as shown in Figure \ref{fig:tu}. In this degraded representation space, owing to loose cluster structures, hard clean samples near the boundaries overlap significantly with noisy samples in distribution \cite{PLC,unicon}. Consequently, relying solely on passive signals such as loss values or confidence is insufficient to provide effective discriminative information, which limits the purity of the selected set and ultimately constrains the upper bound of the performance of the model.

To address this issue, this study is the first to emphasize the importance of active boundary restoration for noise-robust learning. In the LNL domain, existing data augmentation or regularization methods rely primarily on simple feature perturbations \cite{mixup,liu2020early}. We argue that boundaries should be corrected by actively probing inter-class gaps and enhancing representation separability \cite{VOS}. To this end, we propose a novel paradigm based on Hamiltonian dynamics: HamBR.

The core idea of HamBR models the synthesis of virtual outliers as a sampling process based on Markov Chains \cite{neal2011mcmc}. By constructing a potential energy function $U(\mathbf{z})$ based on $K$-Nearest Neighbor ($K$NN) distance, we can quantitatively characterize the probability density topology of samples belonging to the clean manifold in the feature space \cite{energy}. On this basis, we utilize the Spherical Hamiltonian Monte Carlo (Spherical HMC) \cite{lan2014spherical} mechanism to actively probe inter-class ambiguous regions within the representation space and synthesize high-quality virtual outliers. We employ contrastive learning to use these synthesized virtual outliers as negative anchors, constructing geometric regularization constraints together with class centers \cite{supervised}. By minimizing the distance between feature embeddings and class centers while maximizing the discriminative margin with boundary virtual samples, HamBR forces real samples to move from overlapping regions toward class centers, thereby restoring the discriminative sharpness of the decision boundary and achieving effective geometric isolation of noisy signals \cite{deng2019arcface}.

Our main contributions are summarized as follows:
\begin{itemize}
    \item \textbf{Proposed a new perspective of decision boundary restoration:} We reveal that boundary collapse is the key bottleneck limiting the performance of sample selection, realizing a paradigm shift from passive screening to active geometric boundary restoration.
    
    \item \textbf{Designed an active probing mechanism based on Hamiltonian dynamics:} By modeling sampling as a Markov Chain and using an energy topology model based on $K$NN-distance, we efficiently synthesize boundary outliers in the hyperspherical space, which establishes a theoretical foundation for subsequent discriminative enhancement.
    
    \item \textbf{Significantly improved performance and robustness:} HamBR significantly improves the identification accuracy of edge hard samples and the efficiency of training convergence. Experiments show that this method achieves state-of-the-art (SOTA) performance on CIFAR-10/100 and real-world datasets, and demonstrates superior capability for Out-of-Distribution (OOD) sample detection.
\end{itemize}

\begin{figure*}[t]
    \centering
    \includegraphics[width=0.99\textwidth]{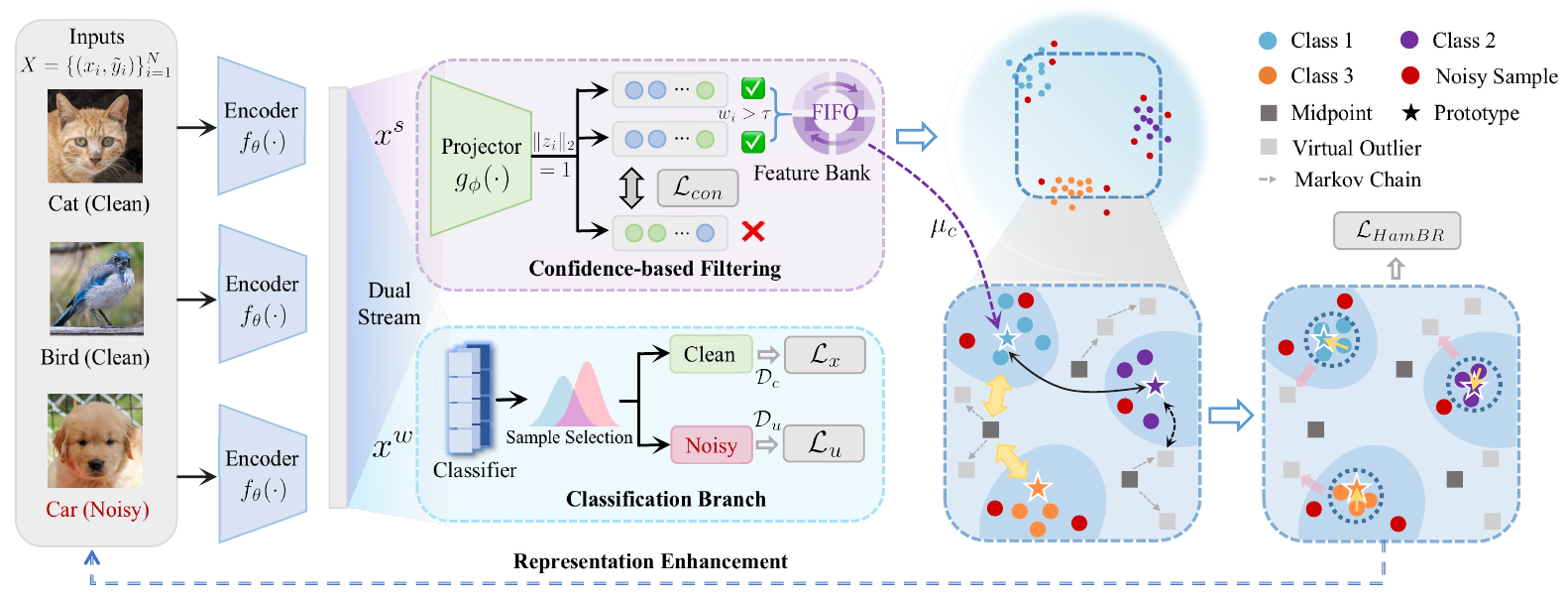} 
    \caption{\textbf{Overview of the HamBR Framework.}The framework comprises a classification branch for sample partitioning and a filtering branch for maintaining a feature bank. The core module implements active geometric regularization ($\mathcal{L}_{\text{HamBR}}$) by synthesizing virtual outliers between class prototypes, which forces clean samples towards prototypes and away from virtual outliers, thereby optimizing the feature manifold structure and enhancing the model's discriminability.}
    \label{fig:1}
\end{figure*}
\section{Related Work}

\subsection{Sample Selection and Semi-Supervised Noisy Learning}
Sample selection is one of the most prominent paradigms for handling noisy labels. Its core assumption relies on the small-loss trick, which suggests that deep networks tend to prioritize the learning of simple and clean samples \cite{arpit2017closer}. Early works, such as Co-teaching and its variants, use dual networks to cross-filter samples and mitigate accumulated errors \cite{yu2019does}. DivideMix \cite{li2020dividemix} extends this idea to a Semi-Supervised Learning (SSL) framework by using a Gaussian Mixture Model (GMM) to dynamically divide data into a labeled set (clean) and an unlabeled set (noisy), and applying the MixMatch strategy \cite{berthelot2019mixmatch} for joint training. Subsequent studies, such as LongReMix \cite{longremix} and PSSCL \cite{zhang2025psscl}, have introduced more refined selection strategies or two-stage training processes on this basis to address the challenges of high noise rates or long-tailed distributions. ProMix \cite{promix} attempts to maximize the utility of clean samples by expanding the base clean set through matched high-confidence selection.

Although these methods have made significant progress, they rely primarily on prediction confidence for selection. When noisy samples are located near decision boundaries or have misleadingly high confidence \cite{northcutt2021confident}, GMMs often struggle to distinguish them \cite{xia2020part}. This results in the erroneous categorization of noise into the clean set, thereby contaminating the training of the model. Unlike these methods that rely solely on statistical selection, our approach introduces an active geometric defense mechanism, which reinforces the ability to identify boundary noise through physical constraints \cite{bouveyron2009robust}.

\subsection{Robust Representation Learning}
In addition to label cleaning, another strategy focuses on learning robust feature representations that are insensitive to label noise \cite{zhong2019graph}. RRL \cite{Sel15} proposes the learning of robust representations by regularizing the geometric structure of the feature space without any label cleaning. UNICON \cite{unicon} notes that traditional sample selection tends to select classes that are easy to learn, which leads to an unbalanced feature space. Therefore, it proposes the use of contrastive learning and Jensen-Shannon divergence to enhance the discriminability of features.
However, existing contrastive learning methods focus mainly on pairwise relationships between samples and lack explicit constraints on the boundaries of the feature manifold \cite{chen2020simple,he2020momentum}. Noisy samples often drift at the edges of class clusters or in inter-class voids; relying solely on the minimization of intra-class distances makes it difficult to isolate them completely. Our work establishes explicit energy barriers in the feature space by introducing virtual outliers as negative pivots, thereby geometrically compressing the distribution within classes and isolating noisy samples.

\subsection{Virtual Outlier Synthesis and Manifold Regularization}
Virtual Outlier Synthesis (VOS) \cite{VOS} has been extensively investigated in OOD detection \cite{hendrycks2016baseline}, with the aim of regularizing decision boundaries by generating virtual samples in low-density regions. However, the direct transfer of traditional VOS techniques, typically based on GANs or simple Gaussian sampling, to LNL faces a dual challenge: excessive computational overhead for generation or the inability to capture complex manifold topologies precisely \cite{goodfellow2014generative,lee2018simple}. To bridge this gap, our proposed HamBR pioneers the use of Spherical Hamiltonian Dynamics in LNL. By leveraging the gradient-guided physical dynamics of HMC, HamBR synthesizes high-quality virtual outliers precisely within inter-class voids without requiring auxiliary generative models \cite{li2025outlier}. This process reverses noise-induced boundary collapse by establishing explicit energy barriers.

\section{Method}
As illustrated in Figure \ref{fig:1}, our proposed framework is designed to synergize sample selection with the reshaping of geometric features, with the active geometric regularization mechanism as its core. By leveraging a feature bank to establish initial class prototypes, we initialize Markov chains from the midpoints of these prototypes to synthesize virtual outliers within low-density inter-class regions, effectively filling the boundary gaps. Subsequently, a geometric regularization loss is applied to move clean samples explicitly toward their respective prototypes, enhancing the compactness within classes while simultaneously pushing them away from the synthesized outliers to increase the separation between classes. This process actively reshapes the feature manifold, yielding more compact class clusters and sharper decision boundaries, thereby establishing a positive feedback loop for continuous representation enhancement. The remainder of this section details the strategy for sample partitioning, the algorithm for virtual outlier synthesis, and the overall optimization objective.

\subsection{Representation Space Initialization}
To construct a robust feature space capable of supporting the synthesis of virtual outliers under noise interference, we first establish high-quality geometric anchors through a dual-network architecture. Specifically, we use the Generalized Cross Entropy (GCE) loss to pre-warm the networks, facilitating the acquisition of a coarse representation space insensitive to noise in the early stages. Subsequently, a GMM is used to fit the distribution of sample loss dynamically. 

To mitigate the stochasticity of single-pass division and construct a high-purity HamBR feature bank, we implement a rigorous spatiotemporal consensus filtering mechanism. We introduce a time window $T_{filter}$ and retain only those samples consistently judged as clean by the GMM within a sliding window of $T_{filter}$ consecutive epochs. This mechanism effectively filters out hard noise samples that reside on the division boundary of the GMM and exhibit unstable predictions. The resulting support set $\mathcal{B} = \Omega_{ref}$ characterizes the core topology of the intra-class distribution precisely, providing a solid geometric foundation for the subsequent energy-based outlier synthesis.
\begin{equation}
\Omega_{ref}^t = \bigcap{\tau=t-T_{filter}+1}^{t} {i \mid P(w_i=1 \mid \mathcal{L}_i^{(\tau)}) > \gamma }.
\end{equation}

\subsection{Boundary Detection via Dissipative Spherical Hamiltonian Dynamics}
To actively probe ambiguous regions between classes within the feature space, different from traditional passive generation methods based on Gaussian sampling, we model the synthesis of virtual outliers as a physical motion process on a Riemannian manifold. We propose a Dissipative Spherical Hamiltonian Dynamics (DSHD) mechanism, which leverages geometric gradient information to guide sampling points toward decision boundaries.

\subsubsection{Construction of Energy Potential Surface}
First, we construct a scalar field $U(\mathbf{z})$ to quantify the likelihood that an arbitrary point $\mathbf{z} \in \mathbb{R}^d$ belongs to the clean data manifold. Given a feature bank $\mathcal{B} = \{(\mathbf{k}_i, w_i)\}_{i=1}^N$ maintained by the confidence-based filtering branch, where $\mathbf{k}_i$ denotes the stored feature vector and $w_i$ represents the corresponding clean probability weight, we define this probability distribution using Weighted Kernel Density Estimation (KDE). For any class $c$, its class-conditional probability density $p(\mathbf{z}|c)$ is modeled as a mixture of von Mises-Fisher (vMF) distributions:
\begin{equation}
p(\mathbf{z}|c) \propto \sum_{(\mathbf{k}_j, w_j) \in \mathcal{B}_c} w_j \cdot \exp\left(\frac{\mathbf{z}^\top \mathbf{k}_j}{\tau}\right),
\end{equation}
where $\tau$ is the temperature coefficient. Based on the relationship between the Boltzmann distribution and energy $p(\mathbf{z}) \propto \exp(-U(\mathbf{z}))$, we define the free energy relative to class $c$ as:
\begin{equation}
E(\mathbf{z}; \mathcal{B}_c) = -\tau \log \sum_{(\mathbf{k}_j, w_j) \in \mathcal{N}_K(\mathbf{z})} w_j \cdot \exp\left(\frac{\mathbf{z}^\top \mathbf{k}_j}{\tau}\right).
\end{equation}

To improve computational efficiency and robustness, we consider only the $K$ nearest neighbors $\mathcal{N}_K(\mathbf{z})$ in the feature bank that are most similar to $\mathbf{z}$. Our goal is to identify low-density regions located at the boundaries of all classes. Therefore, we define the global potential function $U(\mathbf{z})$ as the lower bound approximation of all class-conditional energies. This implies searching for the most probable assigned class; if even the energy of the most probable class is high, it indicates that the point lies OOD or at a boundary:
\begin{equation}
U(\mathbf{z}) = \min_{c \in \mathcal{C}} E(\mathbf{z}; \mathcal{B}_c).
\end{equation}

On this energy surface, class centers correspond to potential energy valleys, while inter-class decision boundaries correspond to saddle points.

\subsubsection{Dissipative Dynamics on Riemannian Manifold}
We employ Hamiltonian dynamics to efficiently sample high-energy boundary samples on the unit hypersphere $\mathbb{S}^{d-1}$. A standard Hamiltonian system consists of position $\mathbf{z}$ and momentum $\mathbf{v}$, with state evolution governed by Hamiltonian equations:
\begin{equation}
\frac{d\mathbf{z}}{dt} = \nabla_{\mathbf{v}} H, \quad \frac{d\mathbf{v}}{dt} = -\nabla_{\mathbf{z}} H,
\end{equation}
where the Hamiltonian $H(\mathbf{z}, \mathbf{v}) = U(\mathbf{z}) + \frac{1}{2}\|\mathbf{v}\|^2$ represents the total system energy. However, standard HMC maintains energy conservation, causing sampling points to oscillate repeatedly on the potential surface rather than settling at the desired boundary regions. To address this, we introduce Riemannian friction and stochastic diffusion into the dynamic system, extending it to underdamped Langevin dynamics on the manifold. The continuous-time Stochastic Differential Equation (SDE) is formulated as:
\begin{equation}
\begin{cases} 
d\mathbf{z}_t = \mathbf{v}_t dt \\ 
d\mathbf{v}_t = \underbrace{-\nabla_{\mathcal{S}} U(\mathbf{z}_t) dt} - \underbrace{\gamma \mathbf{v}_t dt} + \underbrace{\sqrt{2\gamma \beta^{-1}} d\mathbf{W}_t} - \underbrace{\|\mathbf{v}_t\|^2 \mathbf{z}_t dt}.
\end{cases}
\end{equation}

The physical interpretation of this equation is as follows: $\nabla_{\mathcal{S}} U$ denotes the Riemannian gradient along the sphere; $\gamma$ is the friction coefficient used to dissipate kinetic energy, enabling particle settlement; $d\mathbf{W}_t$ represents the Brownian motion term to maintain local exploration capability; and $-\|\mathbf{v}_t\|^2 \mathbf{z}_t$ acts as the centripetal force to strictly constrain particles to the sphere. 
This dynamic mechanism effectively prevents the samples from overshooting into the class centers (low-energy valleys) and constraining them within the boundary regions.

To simulate this process, we adopt a split-step integrator, decomposing the update process into three sub-steps:

\noindent \textbf{1. Momentum Half-step Update:}
Introducing friction decay via the damping factor $\alpha = \exp(-\gamma \epsilon)$ and injecting Gaussian noise $\boldsymbol{\xi}$:
\begin{equation}
\mathbf{v}' \leftarrow \alpha \mathbf{v}_t - \frac{\epsilon}{2} \nabla_{\mathcal{S}} U(\mathbf{z}_t) + \sqrt{1-\alpha^2}\boldsymbol{\xi}.
\end{equation}

\noindent \textbf{2. Geodesic Flow Update:}
Moving particles along the great circle arc of the sphere, strictly enforcing the geometric constraint $\|\mathbf{z}_{t+1}\|=1$:
\begin{equation}
\mathbf{z}_{t+1} \leftarrow \mathbf{z}_t \cos(\|\mathbf{v}'\|\epsilon) + \frac{\mathbf{v}'}{\|\mathbf{v}'\|} \sin(\|\mathbf{v}'\|\epsilon).
\end{equation}

\noindent \textbf{3. Momentum Transport and Correction:}
Translating the momentum vector to the new tangent space $\mathcal{T}_{\mathbf{z}_{t+1}}\mathbb{S}^{d-1}$ and applying the remaining gradient force:
\begin{equation}
\mathbf{v}_{t+1} \leftarrow \mathcal{R}(\mathbf{v}', \mathbf{z}_t \to \mathbf{z}_{t+1}) - \frac{\epsilon}{2} \nabla_{\mathcal{S}} U(\mathbf{z}_{t+1}),
\end{equation}
where $\mathcal{R}$ denotes the vector transport operator. Through this process, the generated set of virtual outliers $\mathcal{V} = \{\mathbf{v}_j\}_{j=1}^M$ outlines the topological structure with the most severe inter-class overlap in the current feature space precisely.

\subsubsection{Algorithm Flow}
To ensure comprehensive exploration of the representation space, we run multiple Markov chains in parallel. The initial state $\mathbf{z}^{(0)}$ of each chain is set to the normalized midpoint of two randomly selected class prototypes. The complete generation process is summarized in Algorithm 1.

\subsection{MLE-based Manifold Regularization}
After obtaining the virtual outliers $\mathcal{V}$, we aim to reshape the feature manifold using geometric constraints. Assuming a clean sample $\mathbf{x}$ and its corresponding class prototype $\boldsymbol{\mu}_c$, we seek to maximize the posterior probability $P(c|\mathbf{x})$ of belonging to the correct class $c$. Under the open-space assumption containing $M$ virtual outliers (acting as negative samples/background class), this probability can be expressed in the form of a Boltzmann distribution:
\begin{equation}
P(c|\mathbf{x}) = \frac{\exp(\text{sim}(\mathbf{x}, \boldsymbol{\mu}_c)/\tau)}{Z(\mathbf{x})},
\end{equation}
where the partition function $Z(\mathbf{x})$ must account for both the positive sample (class prototype) and all potential distractors (i.e., synthesized virtual outliers $\mathbf{v}_j \in \mathcal{V}$):
\begin{equation}
Z(\mathbf{x}) = \exp(\text{sim}(\mathbf{x}, \boldsymbol{\mu}_c)/\tau) + \sum_{j=1}^M \exp(\text{sim}(\mathbf{x}, \mathbf{v}_j)/\tau).
\end{equation}

In this context, we ignore prototypes of other classes because virtual outliers are designed to populate the gaps between the current class and confusing classes, constituting the primary competitors for distinguishing the correct category.
Our optimization objective is to maximize the log-likelihood function $\mathcal{L} = \sum \log P(c|\mathbf{x})$. By taking the negative, we obtain the final \textbf{Geometric Perception Regularization Loss}, which aligns with the form of the InfoNCE loss:
\begin{equation}
\mathcal{L}_{\text{HamBR}} = - \log \frac{\exp(\mathbf{x}^\top \boldsymbol{\mu}_c / \tau)}{\exp(\mathbf{x}^\top \boldsymbol{\mu}_c / \tau) + \sum_{j=1}^M \exp(\mathbf{x}^\top \mathbf{v}_j / \tau)}.
\end{equation}

Maximizing $\mathbf{x}^\top \boldsymbol{\mu}_c$ forces clean samples to collapse toward class centers, enhancing the compactness within classes. Minimizing $\mathbf{x}^\top \mathbf{v}_j$ forces samples away from virtual outliers located at the boundaries. Since $\mathbf{v}_j$ occupy the inter-class gaps, this repulsive force explicitly establishes energy barriers at the decision boundaries, thereby blocking the path for noisy samples to drift toward incorrect classes.

\subsection{Joint Optimization Objective}
To utilize useful information in noisy data fully and achieve robust learning, this new paradigm serves as a plug-and-play geometric defense module that can be integrated into semi-supervised noisy label learning frameworks. Following the mainstream paradigm of sample selection, we use collaborative label correction and consistency regularization strategies for the current labeled set $\mathcal{X}_{t}$ and noisy unlabeled set $\mathcal{U}$. For samples in $\mathcal{X}_{support}$, we combine the clean probability predicted by the GMM with model predictions for label co-correction and apply data augmentation strategies to compute the cross-entropy loss:
\begin{equation}
\mathcal{L}_x = - \frac{1}{|\mathcal{X}'|} \sum_{x_i \in \mathcal{X}'} \sum_{c=1}^C y_i^c \log(p_i^c(\theta)),
\end{equation}
where $\mathcal{X}'$ is the augmented labeled dataset, $y_i$ is the corrected pseudo-label, and $p_i(\theta)$ is the model prediction. For samples in $\mathcal{U}$, we do not discard them directly but use the information of their feature distribution. We generate pseudo-labels by sharpening the predictions of the dual-network and compute the Mean Squared Error (MSE) as a consistency loss to enhance the robustness of the model against perturbations:
\begin{equation}
\mathcal{L}_u = \frac{1}{|\mathcal{U}'|} \sum_{u_i \in \mathcal{U}'} \| \hat{q}_i - p_i(\theta) \|_2^2,
\end{equation}
where $\hat{q}_i$ is the sharpened guess label. Furthermore, to prevent the collapse of the model into a single class, we introduce a regularization term $\mathcal{L}_{reg}$. Thus, the basic semi-supervised noisy learning loss is defined as:
\begin{equation}
\mathcal{L}_{SSL} = \mathcal{L}_x + \lambda_u \mathcal{L}_u + \lambda_{reg} \mathcal{L}_{reg}.
\end{equation}

To further enhance the discriminability of features for unlabeled data, we introduce a self-supervised contrastive loss $\mathcal{L}_{con}$. By pulling together different augmented views of the same image and pushing away views of different images, this loss mitigates the over-reliance of the model on potentially incorrect pseudo-labels:
\begin{equation}
\mathcal{L}_{con} = - \sum_{i \in \mathcal{U}_{batch}} \log \frac{\exp(\text{sim}(\mathbf{z}_i, \mathbf{z}_i') / \tau_{con})}{\sum_{k \neq i} \exp(\text{sim}(\mathbf{z}_i, \mathbf{z}_k) / \tau_{con})},
\end{equation}
where $\mathbf{z}_i, \mathbf{z}_i'$ are features of two augmented views of sample $i$, and $\text{sim}(\cdot)$ denotes cosine similarity. Finally, we integrate the aforementioned perceptual losses with the proposed active geometric defense loss $\mathcal{L}_{HamBR}$:
\begin{equation}
\mathcal{L}_{total} = \mathcal{L}_{SSL}+ \lambda_c \mathcal{L}_{con} + \lambda_{hambr} \mathcal{L}_{HamBR},
\end{equation}
where $\lambda_c$ and $\lambda_{hambr}$ are balancing coefficients. By minimizing $\mathcal{L}_{total}$, this joint optimization ensures that while the model uses label information, it is constrained by a strong geometric inductive bias. Noisy samples, which fail to satisfy compactness constraints and are repelled by virtual samples, are geometrically isolated outside the core manifold, effectively blocking overfitting paths.

\begin{algorithm}[t]
    \caption{Underdamped Langevin Dynamics on Riemannian Manifold}
    \label{alg:hamos_dynamics}
    \SetAlgoLined
    
    \KwIn{Feature Bank $\mathcal{B}$, Initial states $\mathbf{z}^{(0)}$, Step size $\epsilon$, Friction $\gamma$, Rounds $R$, Steps $L$.}
    \KwOut{Virtual Outliers set $\mathcal{V}$.}
    
    Initialize momentum $\mathbf{v}^{(0)} \sim \mathcal{N}(0, \mathbf{I})$\;
    Project $\mathbf{v}^{(0)}$ onto tangent space: $\mathbf{v}^{(0)} \leftarrow \mathbf{v}^{(0)} - (\mathbf{v}^{(0)\top}\mathbf{z}^{(0)})\mathbf{z}^{(0)}$\;
    
    \For{$r \leftarrow 1$ \KwTo $R$}{
        Re-sample momentum noise $\boldsymbol{\xi} \sim \mathcal{N}(0, \mathbf{I})$\;
        \For{$\ell \leftarrow 1$ \KwTo $L$}{
            Compute Energy Gradient $\mathbf{g} = \nabla_{\mathbf{z}} U(\mathbf{z}^{(\ell-1)})$\;
            
            $\mathbf{v}' \leftarrow (1-\gamma\epsilon)\mathbf{v}^{(\ell-1)} - \frac{\epsilon}{2}\text{Proj}_{\mathbf{z}^{(\ell-1)}}(\mathbf{g}) + \sqrt{2\gamma\epsilon}\boldsymbol{\xi}$ \tcp*[r]{Dissipative Update}
            
            $\theta \leftarrow \|\mathbf{v}'\|\epsilon$\;
            $\mathbf{z}^{(\ell)} \leftarrow \mathbf{z}^{(\ell-1)}\cos(\theta) + \frac{\mathbf{v}'}{\|\mathbf{v}'\|}\sin(\theta)$ \tcp*[r]{Geodesic Update}
            
            Rotate $\mathbf{v}'$ to new tangent space of $\mathbf{z}^{(\ell)}$\;
            $\mathbf{v}^{(\ell)} \leftarrow \text{Rotate}(\mathbf{v}') - \frac{\epsilon}{2}\text{Proj}_{\mathbf{z}^{(\ell)}}(\nabla U(\mathbf{z}^{(\ell)}))$ \tcp*[r]{Correction}
        }
    }
    \Return $\mathcal{V} = \{\mathbf{z}^{(R)}\}$\;
\end{algorithm}

\begin{table*}[t]
    \caption{Experimental results on CIFAR-10 dataset.}
    \label{tab:cifar10_results}
    
    \centering
    \begin{tabularx}{\textwidth}{l *{6}{Y}}
        \toprule
        Dataset & \multicolumn{5}{c}{CIFAR-10} & \\ 
        Noise Mode & \multicolumn{4}{c}{Sym.} & \multicolumn{1}{c}{Asym.} & \\ 
        
        Method & 20\% & 50\% & 80\% & 90\% & 40\%  & Avg. \\ 
        \midrule
        Standard CE&73.4&58.9&49.2&30.8&60.9&54.6\\
        GCE&77.2&75.7&66.7&44.1&67.7&66.3\\
        DivideMix & 79.5 & 74.6 & 70.7 & 44.4 & 68.1 & 67.5 \\
        \rowcolor{bg_gray}
        RRL & 88.4 & 83.3 & 74.7 & 45.2 & 76.5 & 73.6 \\
        UNICON & 72.8 & 71.5 & 68.1 & 41.8 & 76.4 & 66.1 \\
        \rowcolor{bg_gray}
        ProMix & 87.4 & 83.7 & 68.7 & 49.3 & 77.0 &  73.2\\
        LongReMix & 81.4 & 79.0 & 67.6 & 37.3 & 67.7 & 66.6 \\
        L2B & 86.7 & 82.1 & 59.4 & 37.0 & 76.7 & 68.4 \\
        PSSCL & 80.7 & 82.3 & 77.6 & 46.1 & 71.0 & 71.5 \\
        \rowcolor{bg_blue}
        \textbf{Ours} & \textbf{88.8} & \textbf{85.1} & \textbf{80.3} & \textbf{62.3} & \textbf{77.9} & \textcolor{rose_red}{\textbf{78.9}} \\

        \bottomrule
    \end{tabularx}
\end{table*}

\section{Experiments}
In this section, we benchmark the proposed approach against a comprehensive set of representative methods, including DivideMix \cite{li2020dividemix}, UNICON \cite{unicon}, LongReMix \cite{longremix}, PSSCL \cite{zhang2025psscl}, and others \cite{promix,li2020dividemix}. We evaluate these methods on two synthetic noisy datasets (i.e., CIFAR-10/100 \cite{cifar}), which incorporate both symmetric and asymmetric label noise types. In addition, we use multiple noisy datasets from the real world. Furthermore, we conduct discussions and ablation studies to elucidate the intrinsic mechanisms of HamBR.
\subsection{Experimental Setup}
 To ensure feasibility and practicality in real-world applications, we constrain the training phase to a restricted setting. Specifically, we use a pre-trained ResNet-18 for the CIFAR datasets, VGG-19 for Animal-10N, and ResNet-50 for Food-101. Optimization is performed using Stochastic Gradient Descent (SGD) with a momentum of 0.9 and a weight decay of $1 \times 10^{-3}$. All models are trained for a total of 30 epochs. Regarding the HamBR dynamics, the Riemannian friction coefficient $\gamma$ is set to 0.95, and the sampling process consists of $R=5$ rounds, with $L=3$ Leapfrog steps executed in each round.

\textbf{CIFAR-10/100:} As widely used benchmarks for synthetic noise, both datasets contain 50,000 training samples and 10,000 test samples. We follow standard data preprocessing protocols in our experiments. To simulate label corruption, we introduce two types of synthetic noise: (1) symmetric noise, which uniformly flips true labels to other classes with a probability $r$; and (2) asymmetric noise, designed to mimic fine-grained real-world confusion. Specifically, this manifests as mappings between semantically similar classes in CIFAR-10 and circular flipping within superclasses in CIFAR-100.

\textbf{Animal-10N \cite{selfie}:} This is a benchmark dataset that contains real-world label noise, primarily composed of five pairs of animal species with highly confusing visual features, with an estimated intrinsic noise rate of approximately 8\%. Given its prominence in LNL research, we follow established conventions and use the VGG-19 network as the backbone architecture for this dataset.

\textbf{Food-101 \cite{food}:} This dataset consists of real-world food images crawled from the web, which contains naturally occurring label noise. To ensure a fair comparison with prior studies, we use a ResNet-50 pre-trained on ImageNet as the feature extractor.

\subsection{Experimental Results on Synthetic Datasets}
We demonstrate the performance of the proposed paradigm under various label noise scenarios. First, using synthetic noisy label datasets, we consider Sym rates of 20\%, 50\%, 80\%, and 90\%, as well as Asym rates of 40\%.

 \textbf{CIFAR-10:} The quantitative results in Table \ref{tab:cifar10_results} demonstrate that the proposed method establishes a new state-of-the-art in the restricted setting with an average accuracy of 78.9\% and a lead of 5.7\% over the nearest competitor ProMix. This performance gap becomes particularly pronounced under extreme noise conditions where standard sample selection strategies typically fail. Specifically in the most challenging settings such as 90\% symmetric noise where strong baselines like UNICON and LongReMix struggle to maintain discriminative power with accuracies of only 41.8\% and 37.3\% respectively our method retains a robust performance of 62.3\%. This represents a significant improvement of over 13\% compared to ProMix which indicates that the proposed dynamic mechanism effectively prevents model overfitting to heavy noise and clarifies decision boundaries when traditional filtering collapses.
\textbf{CIFAR-100:} The quantitative results in Table \ref{tab:cifar100} demonstrate that the proposed method possesses a significant advantage in handling complex and fine-grained category distributions. While previous state-of-the-art methods such as ProMix and L2B underperform in this environment with average accuracies of only 42.9\% and 43.0\% respectively, the proposed paradigm achieves a superior average accuracy of 61.3\%. This advantage remains consistent across different noise intensities. Most notably, in the highly challenging third setting where LongReMix drops to 47.9\%, our method maintains a robust performance of 56.1\% and surpasses both UNICON and RRL. This performance gap highlights a critical limitation of standard sample selection strategies when dealing with high inter-class similarity because statistical filtering often fails to distinguish hard clean samples from noisy ones. In contrast, the new paradigm enables the model to retain high discriminative power even under complex decision boundaries.

\begin{table}[t]
    \centering
    \caption{Experimental results on CIFAR-100 dataset.}
    \label{tab:cifar100}
    
    \begin{tabularx}{\columnwidth}{l *{4}{Y}}
        \toprule
        Dataset & \multicolumn{3}{c}{CIFAR-100} \\ 
        Noise Mode & \multicolumn{2}{c}{Sym.} & {Asym.}  \\ 
        Method & 20\% & 50\% & 40\% &Avg. \\ 
       \midrule
        DivideMix & 61.6 &  47.9& 53.1 & 54.2\\
         \rowcolor{bg_gray}
        RRL & 61.0 &  58.6& 53.5 & 57.7 \\
         \rowcolor{bg_gray}
        UNICON & 60.7 & 57.1 & 54.2 &57.3\\
        ProMix & 45.2 & 42.8& 40.6&42.9\\
        LongReMix & 61.3& 56.9 &47.8& 55.3\\
        L2B & 42.9& 46.1& 39.8&42.9\\
        PSSCL & 56.5 & 45.6& 46.0 &49.4\\
        \rowcolor{bg_blue}
        \textbf{Ours} & \textbf{64.2} & \textbf{63.5} & \textbf{56.1} & \textcolor{rose_red}{\textbf{61.3}}\\
        
        \bottomrule
    \end{tabularx}%
    
\end{table}
\subsection{Experimental Results on Real-World Datasets}

\textbf{Animal-10N:} The quantitative results in Table \ref{tab:animal10n} demonstrate that HamBR achieves a superior average accuracy of 79.68\% on the real-world Animal-10N benchmark, significantly outperforming all state-of-the-art methods. While the performance of most existing approaches stagnates between 71\% and 72\% due to the inherent difficulty in distinguishing hard-to-learn clean samples from actual mislabeled ones, our method surpasses the strongest baseline, ProMix, by a remarkable margin of 3.42\%. This improvement underscores HamBR's exceptional tolerance to instance-dependent noise, as it effectively preserves ambiguous yet discriminative samples while suppressing the interference of erroneous labels.

\textbf{Food-101:} As presented in Table \ref{tab:food101}, on the real-world Food-101 noisy dataset, the proposed method achieves a test accuracy of 72.5\%, outperforming all comparison methods, which generally cluster around the 70\% range. Notably, HamBR yields a performance gain of 1.8\% over PSSCL, the strongest baseline. Given the fine-grained nature and instance-dependent noise inherent in Food-101, this improvement underscores the enhanced robustness of the proposed approach in real-world noisy scenarios.

\begin{table}[t]
    \centering
    \caption{Experimental results on Animal-10N dataset using VGG-19N.}
    \label{tab:animal10n}
    \begin{tabularx}{\columnwidth}{l *{2}{Y}}
        \toprule
    
        Method & Ref. & Test Accuracy  \\ 
       \midrule
        DivideMix & ICLR20&  71.5  \\
        RRL & ICCV21 &   73.8  \\
        UNICON & CVPR22 &  71.1 \\
        \rowcolor{bg_gray}
        ProMix & IJCAI23 & 76.2\\
        LongReMix &PR23 &  71.6 \\
        L2B &CVPR24 & 71.4\\
        PSSCL & PR25 &  72.7 \\
        \rowcolor{bg_blue}
        \textbf{Ours} & \textbf{-} &  \textcolor{rose_red}{\textbf{79.7}}\\
        \bottomrule
    \end{tabularx}%
    
\end{table}

\begin{table}[t]
    \centering
    \caption{Experimental results on Food-101 dataset using pre-trained ResNet-50.}
    \label{tab:food101}
    \begin{tabularx}{\columnwidth}{l *{2}{Y}}
        \toprule
    
        Method & Ref. & Test Accuracy  \\ 
       \midrule
        DivideMix & ICLR20&  67.2  \\
        RRL & ICCV21 &  69.7   \\
        UNICON & CVPR22 & 70.4  \\
        ProMix & IJCAI23 & 70.1\\
        LongReMix &PR23 &  68.7 \\
        L2B &CVPR24 & 68.9\\
        \rowcolor{bg_gray}
        PSSCL & PR25 &  70.7 \\
        \rowcolor{bg_blue}
        \textbf{Ours} & \textbf{-} & \textcolor{rose_red}{\textbf{72.5}}  \\
        \bottomrule
    \end{tabularx}%
    
\end{table}
\begin{figure*}[htbp]
    \centering
    \begin{subfigure}[b]{0.24\linewidth} 
        \centering
        \includegraphics[width=\linewidth]{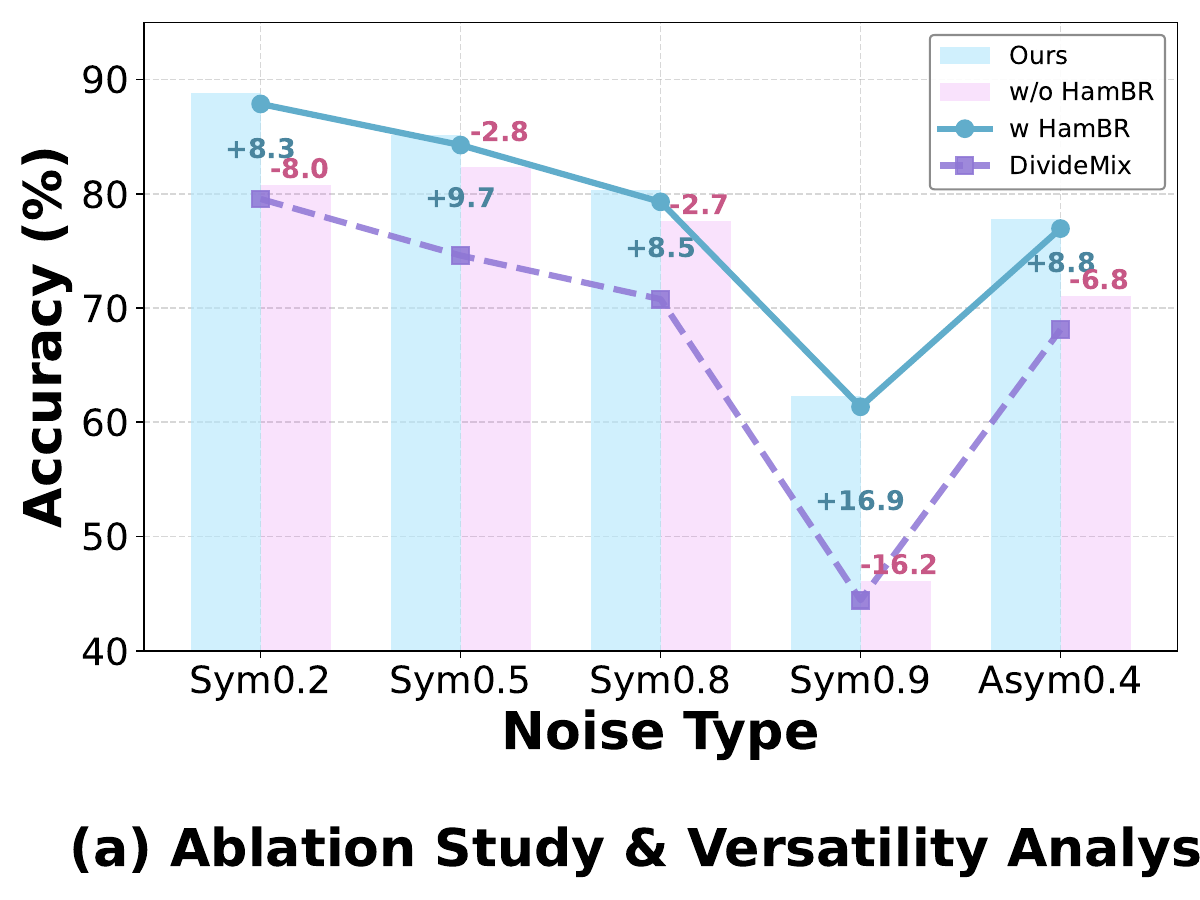}
        \label{fig:sub1}
    \end{subfigure}
    \hfill 
    \begin{subfigure}[b]{0.24\linewidth}
        \centering
        \includegraphics[width=\linewidth]{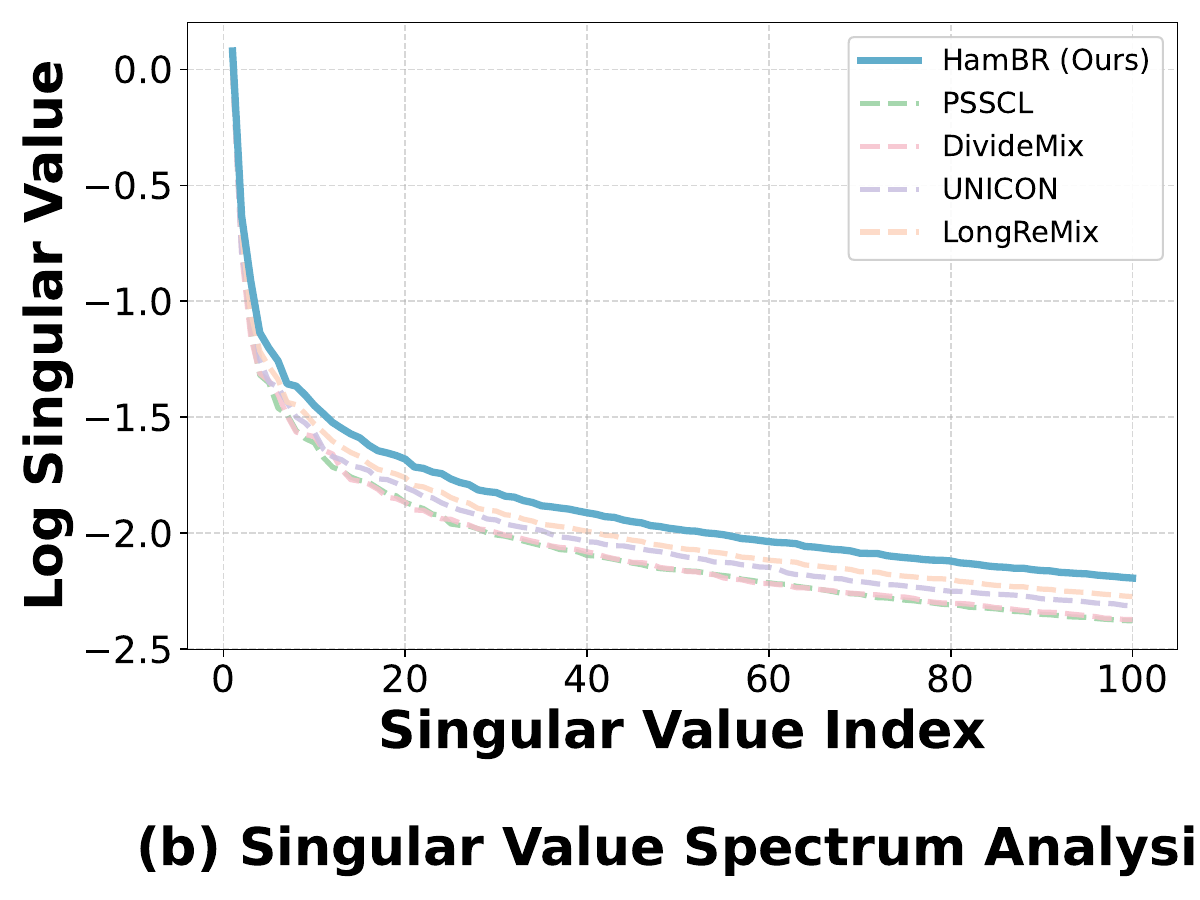}
        \label{fig:sub2}
    \end{subfigure}
    \hfill 
    \begin{subfigure}[b]{0.24\linewidth}
        \centering
        \includegraphics[width=\linewidth]{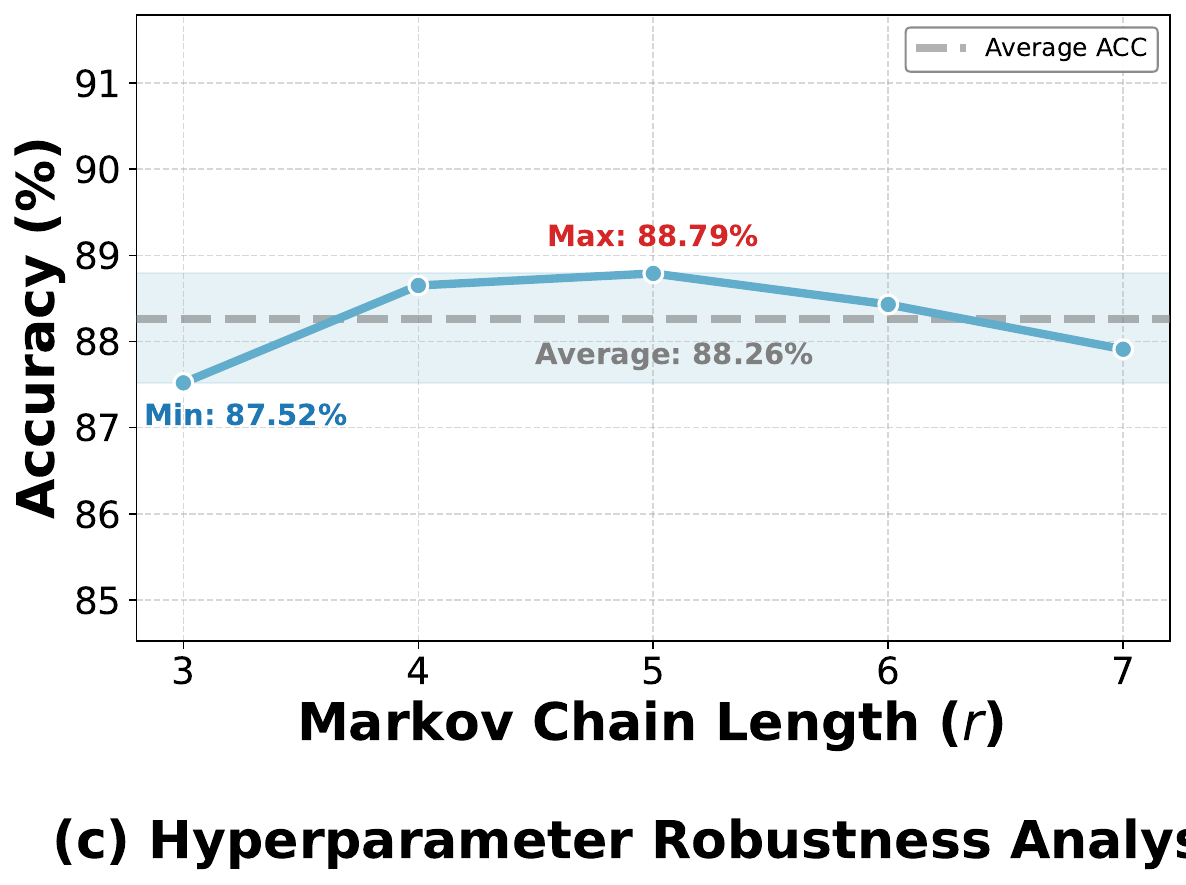}
        \label{fig:sub3}
    \end{subfigure}
    \hfill 
    \begin{subfigure}[b]{0.24\linewidth}
        \centering
        \includegraphics[width=\linewidth]{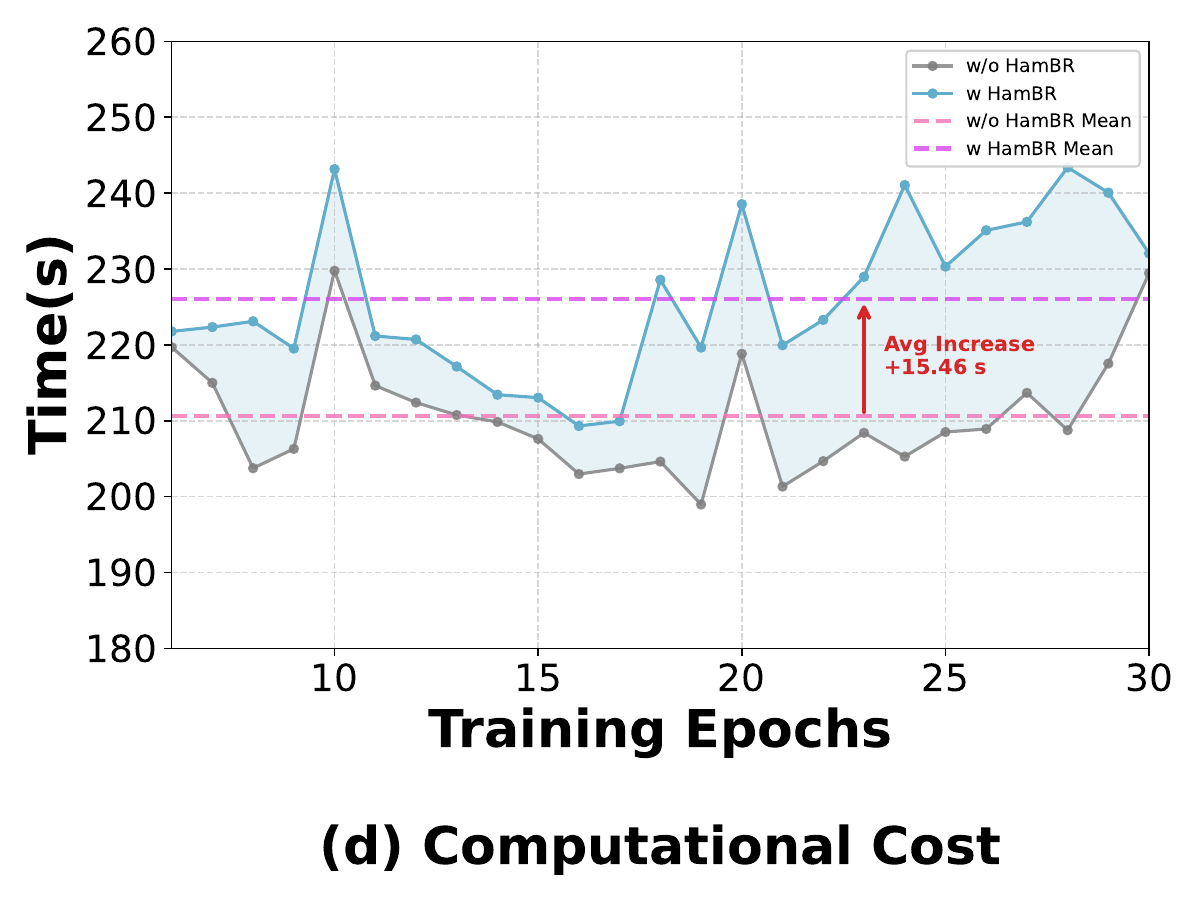} 
        \label{fig:sub4} 
    \end{subfigure}
    \caption{(a) Ablation study of the HamBR module and verification of its versatility within the DivideMix framework. (b) Singular value spectrum analysis of the feature space. (c) Parameter sensitivity analysis on the Markov chain length $r$. (d) Analysis of per-epoch training time costs.}
    \label{fig:main}
\end{figure*}

\subsection{Detailed Analysis}
\subsubsection{Ablation Study of HamBR Module}
As shown in Figure \ref{fig:main}(a), the HamBR module significantly enhances model robustness across all noise configurations, yielding an average performance improvement of approximately 7.32\%. This advantage is particularly pronounced under extreme noise conditions, at a 0.9 symmetric noise ratio, the module substantially boosts accuracy from 46.10\% to 62.34\%, achieving a remarkable margin of 16.24\%. Furthermore, robust gains are consistently observed under asymmetric noise, fully validating HamBR's critical role in preventing boundary collapse within severe noise environments.

\subsubsection{Versatility of HamBR as an Enhancement Module}
To demonstrate the versatility of the HamBR module, we integrated it into the existing DivideMix framework and observed significant performance gains across all noise scenarios. As illustrated in Figure \ref{fig:main}(a), enhancing DivideMix with HamBR improves its performance by approximately 12.46 percentage points on average. This enhancement is particularly critical under extreme noise conditions where the model accuracy surged from 44.43\% to 61.36\% at a 0.9 symmetric noise ratio, achieving a significant margin of 16.93\%. Furthermore, the 8.86\% improvement under asymmetric noise further confirms HamBR as a universal module capable of effectively fortifying existing methods against various types of label corruption. These results indicate that HamBR serves as a general-purpose module to effectively enhance the robustness of existing methods against diverse forms of label corruption.

\subsubsection{Singular Value Spectrum Analysis}
As evidenced by the singular value spectrum analysis in Figure \ref{fig:main}(b), HamBR exhibits a significantly slower decay in singular values compared to other methods, demonstrating that active feature space reshaping effectively mitigates feature collapse. HamBR consistently outperforms baselines across the entire spectral range, particularly in the mid-tail region, indices 20-100. Specifically, at index 40, HamBR maintains a log singular value of approximately $-1.9$, whereas competing methods drop below $-2.0$. This sustained high-amplitude spectrum indicates that HamBR preserves a higher effective rank of the feature matrix. By enforcing geometric constraints, it effectively prevents representation shrinkage into low-dimensional manifolds, thereby retaining richer discriminative structures amidst noise interference. 

\subsubsection{Ablation Study of Moderate Markov Chain Length}
As illustrated in Figure \ref{fig:main}(c), we analyze the impact of the Markov chain length $r$ on classification accuracy. Increasing $r$ from 3 to 5 yields a steady improvement in accuracy from 87.59\% to 88.79\%, suggesting that a moderate number of sampling steps allows the model to more thoroughly explore low-energy and well-structured regions within the representation space. However, when the chain length is further increased to 6 and 7, the accuracy drops to 87.94\% and 88.10\%, respectively, indicating that the performance gains do not persist. This observation implies that excessively long Markov chains may introduce redundant sampling or additional noise, which can adversely affect the learned representations. Overall, the method demonstrates a degree of robustness to variations in chain length, achieving an optimal trade-off between sampling sufficiency and computational efficiency at $r=5$.

\subsubsection{Per-Epoch Training Cost Analysis}
As shown in Figure \ref{fig:main}(d), the training time with HamBR is consistently higher than that without HamBR. This indicates that the additional overhead introduced by HamBR is systematic, arising mainly from virtual sample synthesis and the associated energy-based constraint computation, rather than from occasional fluctuations caused by specific parameter settings. Without HamBR, the training time is roughly between 199 s and 229 s, whereas the version with HamBR ranges from approximately 209 s to 243 s. Overall, although HamBR consistently increases the training time, the magnitude of this increase remains moderate and within a controllable range for most settings, reflecting a stable trade-off between the benefits of enhanced representation structure and computational efficiency.

\subsubsection{Comparison of Different Sampling Strategies}
To rigorously evaluate the contribution of our active sampling mechanism, we benchmark HamBR against three representative sampling baselines: (1) Gaussian Sampling (low-likelihood sampling based on GMM), (2) Mixup Sampling (linear interpolation between clean features), (3) Perturbation Sampling (stochastic Gaussian noise scaled by feature standard deviation). As shown in Table \ref{tab:ablation_sampling}, in the low-noise regime, our method achieves a dominant accuracy of 88.8\%, surpassing the nearest competitor, Mixup, by a substantial margin of 10.2\%. This performance gap underscores the intrinsic limitations of passive augmentation techniques, which fail to actively rectify decision boundaries. The robustness of our framework becomes particularly pronounced under severe noise conditions. As the noise ratio escalates to 0.8 and 0.9, both Gaussian sampling and Mixup exhibit drastic performance deterioration, plummeting to approximately 47\% accuracy. This demonstrates that while stochastic noise injection and linear interpolation struggle to preserve manifold structures amidst overwhelming noise, our active boundary restoration mechanism successfully prevents the model from degenerating into random guessing.

\begin{table}[htbp]
    \centering
    \caption{Comparison of different sampling strategies.}
    \label{tab:ablation_sampling}
    
    \begin{tabularx}{\columnwidth}{l *{5}{Y}}
        \toprule
        Method & Sym0.2 & Sym0.5 & Sym0.8 & Sym0.9 & Avg. \\
        \midrule
        Gaussian & 75.9 & 71.1 & 65.8 & 47.0 & 65.0 \\
        Mixup & 78.6 & 71.2 & 57.0 & 47.1 & 63.5 \\
        Perturbation & 76.4 & 72.0 & 61.4 & 46.9 & 64.2 \\
        \rowcolor{bg_blue}
        \textbf{Ours} & \textbf{88.8} & \textbf{85.1} & \textbf{80.3} & \textbf{62.3} & \textcolor{rose_red}{\textbf{79.1}} \\
        \bottomrule
    \end{tabularx}
\end{table}
\vspace{-5pt}
\subsection{Noise Robustness in OOD Detection}
Following the OpenOOD benchmark protocols, we conducted a comprehensive evaluation of the proposed paradigm's OOD performance across five standard datasets, namely SVHN, LSUN, Places365, Textures, and MNIST. To quantify the detection efficacy for OOD data, we employed two standard metrics: AUROC and FPR95. The results in Table \ref{tab:cifar10_custom} consistently demonstrate that our method achieves SOTA detection performance across all evaluated symmetric noise levels. Notably, at noise ratios of 0.2 and 0.5, the average FPR95 is as low as 18.3\% and 19.6\%, respectively, significantly outperforming existing baselines. These findings provide empirical evidence that the proposed method effectively mitigates boundary collapse, resulting in more compact in-distribution (ID) feature representations and achieving a clear separation from OOD samples.
\ifHamBRIncludeAppendix
\else
Detailed OOD breakdown tables appear in supplementary material omitted from this arXiv preprint.
\fi
\begin{table}[t]
    \centering
    \setlength{\tabcolsep}{3pt}
    \caption{Comparison of OOD detection performance on CIFAR-10.}
    \label{tab:cifar10_custom}
    
    \begin{tabularx}{\columnwidth}{l *{4}{Y}}
        \toprule
        Dataset & \multicolumn{4}{c}{CIFAR-10} \\ 
        Noise Mode & Sym0.2 & Sym0.5 & Sym0.8 & Asym0.4  \\ 
        Method &  \multicolumn{4}{c}{AUROC $\uparrow$/FPR95 $\downarrow$}   \\ 
        \midrule

      
        DivideMix & 69.7/64.1 & 77.2/51.2 & 77.9/67.0 &  49.2/93.1\\
        RRL & 66.3/84.9 &  51.1/98.8& 30.8/99.1 & 41.7/98.0 \\
        
        UNICON & 84.7/38.4 & 92.9/19.9 & 83.9/35.3 &82.5/33.9  \\
        ProMix & 76.6/45.6 & 82.0/36.0 &79.4/40.5  &81.2/39.2  \\
        
        LongReMix & 70.8/64.9 & 68.8/54.7 & 62.6/62.5 &70.5/68.6  \\
        L2B &  70.8/64.9& 75.5/64.9 &68.3/82.8& 78.9/66.4 \\
        
        PSSCL & 93.4/24.0 & 92.3/26.7 & 87.8/31.1 & 87.5/33.8 \\
        \rowcolor{bg_blue}
        \textbf{Ours} & 
        \textbf{96.0/18.3}& 
        \textbf{95.9/19.6} & 
        \textbf{91.0/26.3}& 
        \textbf{93.7/22.3} \\
        \bottomrule
    \end{tabularx}
\end{table}


\section{Conclusion}
In this paper, we address the critical bottleneck of boundary collapse in LNL. Moving beyond passive sample selection mechanisms, which struggle to distinguish between hard clean samples and noisy ones within degenerated feature spaces, we introduce a novel paradigm centered on active boundary rectification. This fundamental shift empowers the model to enhance the separability of learned representations proactively. Our proposed method, HamBR, is grounded in the principles of Hamiltonian dynamics, specifically incorporating a spherical HMC mechanism to actively probe ambiguous inter-class regions and synthesize high-quality virtual outliers. By constructing robust energy barriers through explicit repulsive constraints, these synthesized samples drive real data points toward their respective class centers, effectively disentangling overlapping regions. Consequently, this process restores the discriminative sharpness of decision boundaries, yielding a significantly more robust and noise-resilient feature manifold. Extensive empirical evaluations validate the efficacy of HamBR, which consistently achieves SOTA performance across CIFAR-10/100 and various real-world noise benchmarks, along with enhanced OOD detection capabilities. Its design as a plug-and-play defense module further underscores its versatility, offering a broadly applicable solution for real-world noisy data challenges.


\bibliographystyle{ACM-Reference-Format}
\bibliography{sample-base}


\end{document}